\documentclass[11pt]{article}
\usepackage[utf8]{inputenc}
\usepackage[T1]{fontenc}
\usepackage{amsmath,amssymb}
\usepackage{booktabs}
\usepackage{graphicx}
\usepackage{url}
\usepackage[hidelinks]{hyperref}
\usepackage{geometry}
\usepackage{pgfplots}
\pgfplotsset{compat=1.17}
\usepackage{algorithm}
\usepackage{algpseudocode}
\title{SchemaRouter: Field-Aware Tool Routing for\\ Efficient Heterogeneous Agentic RAG}
\author{Yong-eun Cho \\
  KailosLab \\
  Seoul, Republic of Korea \\
  \texttt{kevin@kailoslab.com}}
\date{\today}

\begin{document}
\maketitle

\begin{abstract}
Heterogeneous agentic retrieval-augmented generation (RAG) systems increasingly orchestrate many
external APIs, internal databases, vector stores, and graph stores across domains as varied as commerce,
media, enterprise, and science. Exposing every tool description to an LLM agent, or retrieving tools by
vector similarity, leads to two failure modes with real cost:
\emph{over-fetching} (selecting unneeded tools/fields inflates retrieval payload, tokens, and latency)
and \emph{under-fetching} (missing required fields yields unanswerable retrievals). We present
\textbf{SchemaRouter}, a lightweight routing layer that models tools, endpoints, parameters, response
fields, domain concepts, units, provenance, and license as a \emph{schema graph}, and emits an
executable tool plan specifying exactly which tools and response fields to fetch. A small LLM step
extracts intent, domain concepts, and source constraints; field selection is then \emph{deterministic}
over the graph (intent-group projection $\cup$ concept--field matching bridged by a domain-concept alias
layer), guaranteeing valid, minimal, provenance-annotated plans. On a materials-science benchmark
($N{=}110$) we evaluate not only planning accuracy but downstream RAG effects: retrieved-context tokens,
end-to-end latency, final-answer accuracy, and \emph{answer trustworthiness} (provenance/license
attribution). SchemaRouter attains
answer accuracy of 0.71---matching fetch-everything within CI and above prompt-all's 0.66 (CIs
overlap)---at near-minimal retrieved context (227 tokens vs.\ 2066 for fetch-everything) and
$\sim$$2.7\times$ lower end-to-end latency than prompt-all (its routing prompt is schema-independent), with
the best tool-exact rate (0.93) and parameter validity (1.0). It further grounds schema-sourced provenance
and license in 62\% of answers---including per-database license strings a reader LLM cannot produce from
parametric knowledge---versus $\approx$0\% for all baselines. We further report a methodological finding:
optimizing planning-level \emph{over-fetch} is counterproductive. Minimizing selected-field count cuts
downstream accuracy to 0.56 at negligible token savings; a recall-preserving projection recovers top
accuracy. Retrieved-context tokens and downstream accuracy---not field-count parsimony---are the objectives
that matter. We frame SchemaRouter's contribution as \emph{efficiency, schema-independent scaling, and
verifiable provenance/license-grounded answers at competitive accuracy}.
\end{abstract}

\section{Introduction}
\label{sec:intro}
Modern agentic RAG systems for science route queries across many heterogeneous data sources---materials
property databases, chemical databases, literature APIs, internal experiment stores. As the tool
inventory grows, the agent must decide not merely \emph{which tool} to call, but \emph{which endpoint},
\emph{which parameters}, and \emph{which response fields} it actually needs. Two common strategies scale
poorly. Placing all tool/field descriptions in the prompt (\emph{prompt-all}) is accurate with a strong
LLM but injects the entire schema into every routing call, inflating cost and latency and eventually
exceeding context limits. Retrieving tools by embedding similarity (\emph{vector-only}) does not reliably
yield \emph{executable} plans with valid parameters and fields.

The motivation for this work is not accuracy in isolation but the \emph{downstream cost of imprecise
retrieval}. Fetching tools and fields freely, or fetching everything, (i) slows retrieval, (ii) returns
unnecessary data, (iii) enlarges the context the reader LLM must process, and (iv) risks diluting the
answer with irrelevant information. A middle layer that fetches only what is needed should save tokens,
searches, and time while preserving answer quality.

We propose \textbf{SchemaRouter}, a schema-graph routing layer that produces minimal, executable,
provenance-annotated tool plans. The method is \emph{domain-independent by construction}: it assumes only
a tool ecosystem whose endpoints expose typed parameters and response fields with attachable metadata
(units, source, license)---a structural pattern present in commerce, media, enterprise, and scientific APIs
alike. In this paper we instantiate and evaluate it on a materials-science tool set, where provenance and
licensing are first-class concerns; empirical cross-domain evaluation is left to future work. We nonetheless
note that license/attribution metadata---our strongest answer-level result---is a cross-domain concern
(media rights, data-use terms), suggesting the mechanism is useful beyond scientific tools.

Our contributions:
\begin{enumerate}
  \item A \emph{field-aware} formulation of tool routing: selection at the response-field granularity,
        not just tool granularity, with a domain-concept alias layer that bridges the vocabulary gap
        between natural-language queries and canonical schema field names.
  \item A \emph{downstream-centric} evaluation: beyond planning metrics, we measure retrieved-context
        tokens, end-to-end latency, and final-answer accuracy under a controlled fixture protocol.
  \item An honest characterization with a methodological finding. Recall-preserving SchemaRouter reaches
        the top group in answer accuracy at near-minimal retrieved-context cost, and grounds verifiable
        provenance and license that a reader LLM cannot recover from parametric knowledge.
        Over-fetching (fetch-all) wastes cost without improving accuracy under a strong reader. Conversely,
        minimizing planning-level over-fetch is counterproductive: it sacrifices downstream accuracy for a
        field-count metric that barely affects token cost.
\end{enumerate}

\section{Related Work}
\label{sec:related}
\textbf{Tool-use LLMs.} Toolformer~\cite{schick2023toolformer}, Gorilla/APIBench~\cite{patil2023gorilla},
API-Bank~\cite{li2023apibank}, and ToolLLM~\cite{qin2023toolllm} study when and how LLMs call APIs and
evaluate call correctness and hallucination. We differ by explicitly evaluating \emph{endpoint,
parameter, and response-field} selection after tool selection, and by measuring the downstream RAG
consequences of that selection.
\textbf{GraphRAG and graph-based tool retrieval.} GraphRAG~\cite{edge2024graphrag} builds knowledge graphs
over the corpus for query-time augmentation; Graph RAG-Tool Fusion~\cite{graphragtoolfusion2025} and
Agent-as-a-Graph~\cite{agentasagraph2025} model tool dependencies as a graph for retrieval. SchemaRouter
graphs the \emph{tool/API schema itself} (fields, units, provenance, license) to produce executable,
minimal plans, rather than retrieving knowledge or ranking tools by relevance. In the production RAG
system from which our benchmark's tool inventory is drawn, the schema graph is maintained in an embedded
property-graph store~\cite{kuzu};\footnote{The production RAG agent is a proprietary internal system and is
not released with this paper. To keep the evaluation harness self-contained and reproducible, we
serialize a frozen JSON export of the graph and read it directly---the released harness therefore does
not depend on Kuzu at runtime. All schema signals used by SchemaRouter (nodes, edges, aliases, provenance,
license) are preserved in the export; see Appendix~\ref{app:ablation} for the isolated contribution of
each signal.} the released evaluation harness reads a frozen JSON export of that graph so that all
routers operate over an identical, versioned schema without a runtime graph-DB dependency.
\textbf{Schema linking and structured tool calling.} Field selection is a sibling problem to
\emph{schema linking} in text-to-SQL, where models align natural-language mentions with columns of a
target schema. Early work coupled linking to the parser via relation-aware
attention~\cite{wang2020ratsql}; more recent systems decouple linking from downstream generation with
either an element-level ranker~\cite{li2023resdsql} or a decomposed prompting
pipeline~\cite{pourreza2023dinsql}. SchemaRouter transports this idea from a single-database query
compiler to a multi-tool RAG router: it generalizes the target from columns of one schema to fields of
many heterogeneous APIs, uses a deterministic (non-LLM) selector once concepts are extracted, and adds
a metadata layer (unit, provenance, license) absent in SQL settings. On the tool-planning side,
model-side function calling and task decomposition frameworks~\cite{shen2023hugginggpt} let an LLM emit
structured tool calls end-to-end. These are complementary to, not competing with, a schema router:
SchemaRouter is the layer that decides \emph{which} tools and \emph{which} fields the caller (LLM or
otherwise) should target, and can serve as a retrieval front-end for a function-calling agent that then
executes the plan.
\textbf{Scientific RAG / materials informatics.} Our benchmark targets materials-science APIs, where
units, provenance (calculated vs.\ experimental vs.\ literature), and license constraints are
first-class.

\section{Problem Definition}
\label{sec:problem}
Given a natural-language query $q$ and an inventory of tools $\mathcal{T}$, each tool $t$ exposing an
endpoint, parameters $P_t$, and response fields $F_t$ (with per-field unit, provenance, and license
metadata), a router must output a \emph{tool plan}: a set of $(t, \text{params}, \text{fields}\subseteq
F_t)$ tuples. We evaluate the plan directly (planning-level), and then its downstream effect: the plan
induces a retrieval whose result is passed to a reader LLM that must answer $q$. Let $G(q)$ be the gold
plan; we measure tool/field precision--recall--$F_1$, parameter validity, executable-plan rate,
over-fetch ratio, and metadata completeness, plus downstream answer accuracy, retrieved-context tokens,
and latency.

\section{Method: SchemaRouter}
\label{sec:method}
\textbf{Schema graph.} Nodes: \texttt{Tool}, \texttt{Endpoint}, \texttt{Parameter},
\texttt{ResponseField}, \texttt{DomainConcept}, \texttt{Unit}, \texttt{SourceType},
\texttt{LicensePolicy}, \texttt{FieldGroup}, \texttt{Intent}. Edges connect endpoints to their
parameters and fields, fields to units/source-type/concepts, and concepts to canonical fields
(\texttt{MAPS\_TO\_CONCEPT}/\texttt{ALIASED\_AS}). The graph is exported once from the target system's API
registry and frozen for reproducibility.

\textbf{Routing.} (1) A small LLM step extracts \{intent, domain concepts, parameter values, source
constraint, explicitly named tools\}. (2) Candidate tools are selected by intent domain and required
field groups; explicitly named tools force a selection; a source constraint filters by provenance
(experimental/calculated/literature). When the query is ambiguous and no tool is named, candidates are
ranked by concept-match score (exact field match weighted above substring), then schema priority, then
fewest total fields---never rewarding breadth. (3) \emph{Field projection is deterministic}: identifiers
are always included; concept--field matches are added (normalized, with a domain-concept alias layer, so
``bulk modulus''$\to$\texttt{k\_vrh}, ``magnetic moment''$\to$\texttt{total\_magnetization}); the intent's
field groups are added as a recall safety net. (4) Unit/provenance/license requirements are attached from
the graph.

This design makes plans valid by construction (fields drawn from $F_t$, parameters from $P_t$), minimal
(no whole-catalog dumping), and metadata-complete, using only a small routing prompt.
Algorithm~\ref{alg:sr} summarizes the procedure.

\begin{algorithm}[t]
\caption{SchemaRouter (per query)}\label{alg:sr}
\begin{algorithmic}[1]
\Require query $q$; schema graph $G$; intent registry $\mathcal{I}$
\Ensure tool plan $\pi$ (list of $(t, \mathit{params}, \mathit{fields}, \mathit{meta})$)
\State $(i, C, P, S, T) \gets \textsc{Understand}(q)$
  \Comment{LLM: intent, concepts, params, source-constraint, named tools}
\If{$T \neq \emptyset$}
  \State $\mathcal{T}_\text{cand} \gets \{ \textsc{Resolve}(G, t) : t \in T \}$
    \Comment{explicit tools override}
\Else
  \State $\mathcal{T}_\text{cand} \gets \{ t \in G.\text{tools} : \text{domain}(t) \in \mathcal{I}[i].\text{domains} \land \text{groups}(t) \cap \mathcal{I}[i].\text{required} \neq \emptyset \}$
  \If{$S \neq \bot$}
    \State $\mathcal{T}_\text{cand} \gets \{ t \in \mathcal{T}_\text{cand} : \text{sourceType}(t) = S \}$
  \EndIf
  \State $\mathcal{T}_\text{cand} \gets \textsc{Rank}(\mathcal{T}_\text{cand}, C)$
    \Comment{concept-match score; web deprioritized; priority; fewest fields}
\EndIf
\State $\pi \gets \emptyset$
\ForAll{$t \in \mathcal{T}_\text{cand}$}
  \State $F \gets \textsc{Identifiers}(t)$
  \ForAll{$c \in C$}
    \State $F \gets F \cup \textsc{ConceptMatch}(G, t, c)$
      \Comment{normalized name + alias layer}
  \EndFor
  \State $F \gets F \cup \bigcup_{g \in \mathcal{I}[i].\text{required}} \textsc{FieldsOf}(G, t, g)$
    \Comment{recall safety net}
  \State $\mathit{meta} \gets \textsc{AttachMeta}(G, t, F)$
    \Comment{unit, source\_type, license}
  \State $\pi.\text{append}((t,\ P|_{\text{params}(t)},\ F,\ \mathit{meta}))$
\EndFor
\State \Return $\pi$
\end{algorithmic}
\end{algorithm}

\section{Benchmark}
\label{sec:benchmark}
The frozen tool inventory contains 19 heterogeneous tools---materials-property databases (Materials
Project, OQMD, JARVIS, NOMAD, COD, and others), chemistry (PubChem), literature APIs (arXiv, Crossref,
OpenAlex, PubMed, Europe PMC), a business-disclosure source (DART), and web/Q\&A fallbacks---exposing 166
response fields in total. We build a benchmark of $N{=}110$ queries across nine categories:
single-tool property lookup (36), chemical lookup (12), provenance-sensitive (12), multi-field property
lookup (10), literature search (10), cross-source comparison (8), unit-sensitive (8), negative queries
with no suitable tool (8), and ambiguous single-entity queries (6). The category mix is unbalanced
(single-tool dominant); we report both micro- and category-averaged results and caution that macro
averages weight sparser categories equally.

Gold plans are grounded in the frozen schema (valid tools/fields by construction). For the downstream
study we use frozen fixtures: each tool record assigns deterministic ground-truth values to every field;
requested fields carry the answer, other fields act as distractors. Answer correctness is judged by an LLM
judge given the gold values (robust to formatting); negative queries require the reader to decline.

\section{Experiments}
\label{sec:experiments}
\textbf{Setup.} Self-hosted vLLM (Qwen-3.5, 122B, MoE), temperature 0, reasoning disabled. Baselines:
prompt-all, vector-only (TEI embeddings), rule-based, and fetch-all (all tools, all fields). Retrieval is
simulated over fixtures; context tokens and latency are measured from the reader call's usage.

\begin{table}[t]\centering
\caption{Planning metrics ($N{=}110$). Over-fetch lower is better.}
\begin{tabular}{lccccc}
\toprule
Router & Tool Exact & Field $F_1$ [95\% CI] & Param Valid.\ & Over-fetch$\downarrow$ & Exec.\ Rate\\
\midrule
prompt-all  & 0.80 & 0.79 [0.73,0.84] & 1.00 & 0.22 & 0.90\\
vector-only  & 0.86 & 0.61 [0.56,0.65] & 0.46 & 0.27 & 0.30\\
rule-based   & 0.83 & 0.72 [0.65,0.78] & 1.00 & 0.12 & 0.90\\
\textbf{SchemaRouter} & \textbf{0.93} & 0.59 [0.53,0.65] & 1.00 & 0.46 & \textbf{0.93}\\
\bottomrule
\end{tabular}
\end{table}
\noindent SchemaRouter's higher planning over-fetch (0.46) is \emph{deliberate}: it selects intent-relevant
field \emph{groups} for downstream recall, not field-count-minimal plans. As \S\ref{sec:oppoint} shows,
this raises downstream accuracy from 0.56 to 0.71 at only $188{\to}227$ retrieved-context tokens---i.e.\
the over-fetch \emph{ratio} is a misleading proxy; actual token cost stays near-minimal.

\begin{table}[t]\centering
\caption{Downstream RAG ($N{=}110$, identical LLM judge, bootstrap 95\% CI). E2E = routing + answer
latency. Answer accuracy tracks field recall; top-4 CIs overlap.}
\begin{tabular}{lccccc}
\toprule
Router & Answer Acc.\ [95\% CI] & Field Recall & Ctx Tokens & E2E (s)\\
\midrule
\textbf{SchemaRouter} & \textbf{0.71} [0.63,0.79] & 0.86 & \textbf{227} & \textbf{5.5}\\
no-field-projection & 0.71 [0.62,0.79] & 0.76 & 321 & 6.2\\
fetch-all    & 0.70 [0.62,0.78] & 1.00 & 2066 & 4.9\\
prompt-all  & 0.66 [0.57,0.75] & 0.84 & 207  & 15.1\\
rule-based   & 0.55 [0.45,0.64] & 0.69 & 176  & 1.5\\
vector-only  & 0.33 [0.25,0.42] & 0.55 & 185  & 11.9\\
\bottomrule
\end{tabular}
\end{table}
\noindent\emph{no-field-projection} = SchemaRouter's tool selection with \emph{all} fields (ablation);
its parity with SchemaRouter confirms recall-preserving field selection captures the needed fields at
lower token cost (227 vs.\ 321).

\begin{table}[t]\centering
\caption{Answer trustworthiness ($N{=}110$, deterministic string check, non-negative queries). Schema-only
license strings (e.g.\ per-DB policies) cannot be produced from the reader's parametric knowledge.}
\begin{tabular}{lccc}
\toprule
Router & Schema-only license cite & License mention & Source-type stated\\
\midrule
fetch-all    & 0.01 & 0.42 & 0.56\\
prompt-all  & 0.00 & 0.59 & 0.62\\
rule-based   & 0.00 & 0.48 & 0.59\\
\textbf{SchemaRouter} & \textbf{0.62} & \textbf{0.94} & \textbf{0.94}\\
\bottomrule
\end{tabular}
\end{table}

\textbf{Findings.}
\begin{itemize}
  \item \emph{Over-fetch wastes cost.} Fetch-all matches prompt-all accuracy within CI while using
        $\sim$$9\times$ the retrieved-context tokens; on ambiguous multi-source queries it further pollutes
        answers with conflicting values.
  \item \emph{Under-fetch harms.} Vector-only prunes and executes poorly; it is significantly worst
        (0.33, non-overlapping CI).
  \item \emph{Downstream accuracy is recall-bound.} Accuracy tracks field recall across routers. With
        recall-preserving field selection (\S\ref{sec:oppoint}) SchemaRouter reaches \textbf{0.71}, tied
        with the highest-recall routers and above prompt-all (0.66; CIs overlap), at 227 retrieved-context
        tokens ($9\times$ fewer than fetch-all) and $\sim$$2.7\times$ lower E2E than prompt-all.
  \item \emph{Trustworthiness is the answer-level advantage.} SchemaRouter grounds provenance
        and per-database license in 62\% of answers via schema-sourced strings that baselines produce at
        $\approx$0\%---critical for verifiable answering (scientific provenance; media/data-use licensing).
\end{itemize}

\textbf{Scalability of routing cost.} The prompt-all baseline places the full tool/field schema in
\emph{every} routing call: measured on our 19-tool/166-field inventory the schema segment is 5{,}042
characters ($\approx$1{,}260 tokens at 4 char/token) per query, growing at $\approx$66 tokens per
additional tool ($\approx$6.6k tokens at 100 tools, $\approx$33k at 500 tools) until it exceeds practical
context budgets. SchemaRouter's routing prompt is schema-\emph{independent}: a fixed instruction listing
the intent registry (1{,}432 characters, $\approx$360 tokens) plus the query. Growth in the routing prompt
is proportional to the number of \emph{intent categories} (not tools) and is negligible in practice.
This mechanism underlies the observed $\sim$$2.7\times$ E2E gap at 19 tools (Table~2); the extrapolation
beyond 19 tools is analytical, not measured, and we flag it as such.

\emph{Could prompt caching close the gap?} Prefix / KV-cache reuse
can amortize a static schema segment across queries and reduce prompt-all's marginal cost for a
\emph{fixed} schema. Two caveats: (i) the schema segment is not static in deployments where the tool
inventory or per-tool descriptions evolve (typical for internal RAG systems where tools are added,
deprecated, or re-permissioned); every change invalidates the prefix. (ii) Even with a perfect cache,
prompt-all still pays the \emph{decode-time attention} cost over the full schema, so latency scales with
inventory size even when input-side cost does not. SchemaRouter is orthogonal to prefix caching: any cache
strategy that helps prompt-all also helps SchemaRouter's much smaller instruction prefix, so the relative
gap is preserved.

\textbf{Is the trustworthiness comparison fair?} Each router's reader receives exactly the metadata its
\emph{own} plan carries. Baseline plans carry none because they do not model provenance/license;
SchemaRouter's do, sourced from the graph. One could also inject metadata into a prompt-all reader---but
that presupposes a source of per-database license/provenance, which is precisely what the schema graph
supplies. The gap thus reflects a genuine capability difference, not a context-construction artifact: the
$\approx$0\% schema-only-license rate for baselines confirms a reader LLM cannot recover these strings
from parametric knowledge.

\textbf{On the LLM judge.} Accuracy is judged by the \emph{same} Qwen-3.5-122B model that generates the
answers, prompted with a stricter rubric (see released harness); this is a same-family judge and we treat
its absolute numbers with caution. Two mitigations: (i) the trustworthiness axis (Table~4) does not use
the judge---it is a deterministic regex check for schema-only license/provenance strings, so the
$\approx$0\% vs.\ 62\% gap is judge-independent; (ii) we conducted a hybrid spot review on a stratified
sample ($n{=}40$) in which an independent LLM (Claude, different family) read the raw answers and
qualitatively confirmed the direction of the trustworthiness result. A fully blinded cross-family judge
over the entire benchmark remains future work.

\textbf{Ablation.} Table~\ref{tab:ablation} isolates each schema signal.
Removing field projection (dump all fields of the selected tools) more than doubles over-fetch and halves
field~$F_1$; removing the provenance layer drops provenance/license completeness to zero; removing
source-type routing degrades tool accuracy on provenance-sensitive queries. A naive
``concept$\to$whole~group'' expansion crashed planning precision, whereas the surgical domain-concept alias
layer raised recall \emph{and} precision, isolating the alias layer as the mechanism that resolves the
query--schema vocabulary gap. Figure~\ref{fig:ablation} visualizes the joint effect on $F_1$, over-fetch,
and provenance completeness.

\subsection{Operating point: planning parsimony vs.\ downstream accuracy}\label{sec:oppoint}
SchemaRouter's field selection exposes a precision/recall knob. The \emph{minimal-field} point minimizes
planning over-fetch (ratio 0.16, field-$F_1$ 0.75) but cuts downstream accuracy to 0.56; the
\emph{recall-preserving} point (adding intent-relevant field groups) raises over-fetch to 0.46 yet lifts
downstream accuracy to 0.71---at only $188{\to}227$ retrieved-context tokens. The two points share
identical tool selection. This exposes a tension: \textbf{minimizing the planning over-fetch \emph{ratio}
is the wrong objective}; it trades away downstream answer accuracy for a field-count metric that barely
affects real token cost. We adopt the recall-preserving point and argue that retrieved-context tokens plus
downstream accuracy, not field-count parsimony, are the objectives practitioners should optimize.

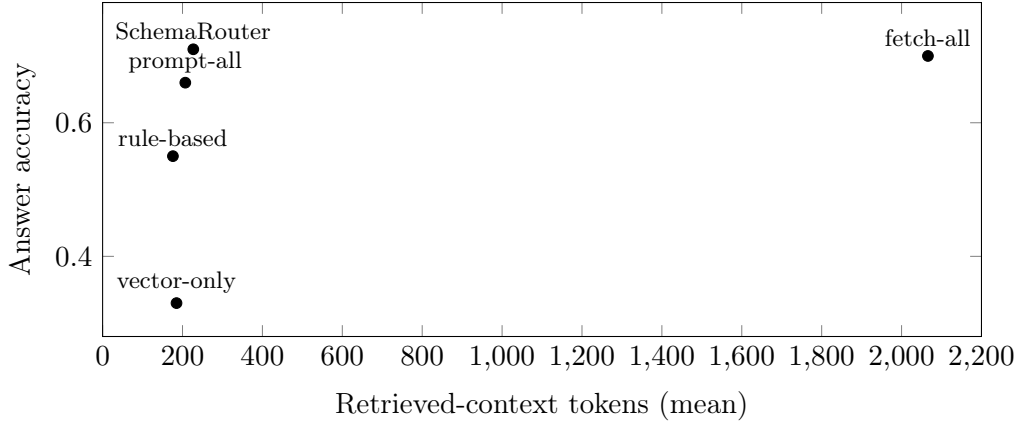
\begin{figure}[t]\centering
\begin{tikzpicture}
\begin{axis}[width=0.8\linewidth, height=6cm, xlabel={Retrieved-context tokens (mean)},
  ylabel={Answer accuracy}, xmin=0, xmax=2200, ymin=0.28, ymax=0.78,
  nodes near coords, every node near coord/.append style={font=\footnotesize, anchor=south},
  scatter/classes={a={mark=*,draw=black}}]
\addplot[scatter, only marks, point meta=explicit symbolic,
  visualization depends on={value \thisrow{lbl}\as\lbl}, nodes near coords={\lbl}]
  table[meta=lbl]{
  x     y      lbl
  227   0.71   SchemaRouter
  2066  0.70   fetch-all
  207   0.66   prompt-all
  176   0.55   rule-based
  185   0.33   vector-only
  };
\end{axis}
\end{tikzpicture}
\caption{Answer accuracy vs.\ retrieved-context tokens ($N{=}110$). SchemaRouter reaches top accuracy at
near-minimal context; fetch-all matches accuracy at $9\times$ the tokens; under-fetching routers collapse.}
\end{figure}

\begin{figure}[t]\centering
\begin{tikzpicture}
\begin{axis}[width=0.8\linewidth, height=4.5cm, ybar, bar width=14pt, ymin=0, ymax=0.7,
  ylabel={Schema-only license cite}, symbolic x coords={fetch-all,prompt-all,rule-based,SchemaRouter},
  xtick=data, nodes near coords, every node near coord/.append style={font=\footnotesize},
  x tick label style={font=\footnotesize}]
\addplot coordinates {(fetch-all,0.01) (prompt-all,0.00) (rule-based,0.00) (SchemaRouter,0.62)};
\end{axis}
\end{tikzpicture}
\caption{Fraction of answers citing schema-only license strings (unavailable from parametric knowledge).
Only SchemaRouter propagates them, from the schema graph.}
\end{figure}
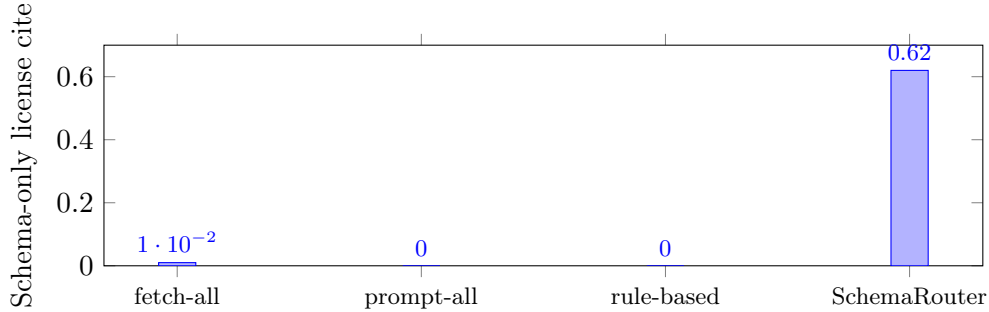

\section{Discussion and Error Analysis}
\label{sec:discussion}
SchemaRouter's residual answer errors decompose into two classes: tool-selection misses (ambiguous
entities routed to a neighboring domain) and reader errors. The field-recall misses that once dominated
this decomposition are resolved by the alias layer plus the recall-preserving intent-group projection
(\S\ref{sec:oppoint}). Under a strong reader, precise (recall-preserving) and bloated retrievals reach the
same top accuracy, so over-fetching buys nothing but cost; the informative axes become tokens, latency,
scalability, and metadata fidelity---on all of which SchemaRouter leads or ties. This equivalence may not
hold for weaker readers, where the accuracy cost of bloat (context pollution) that our motivation
anticipates could re-emerge.

\section{Threats to Validity and Limitations}
\label{sec:threats}
\textbf{Statistical.} $N{=}110$; bootstrap 95\% CIs overlap for the top routers, so we make \emph{no}
answer-accuracy superiority claim---only efficiency (non-overlapping on tokens/latency) and
trustworthiness (non-overlapping, $\approx$0\% vs 62\%). Results are a single run; vLLM at temperature~0
still exhibits scheduler nondeterminism, so we avoid fine-grained accuracy orderings. Multi-seed and
multi-model replication remain future work.
\textbf{Reader strength.} A strong 122B reader resists distraction, so over-fetching \emph{wastes cost}
without measurably \emph{degrading} accuracy (except on ambiguous multi-source queries); weaker readers or
prose retrieval may surface the degradation our motivation anticipates.
\textbf{Fixtures.} Downstream retrieval is simulated with frozen, structured (key=value) fixtures with
deterministic ground-truth values; this controls confounds and enables exact judging but understates the
context pollution of real, unstructured document retrieval. Absolute accuracy is therefore
setup-specific; cross-router \emph{comparisons} under the identical protocol are the intended reading.
\textbf{Benchmark.} Gold plans are author-constructed and partly template-generated (materials-science
centric, single annotator, no inter-annotator agreement); template uniformity makes the set larger but
easier, so numbers are not comparable across benchmark versions.
\textbf{Latency.} End-to-end latencies are measured on a shared, loaded self-hosted server and are
indicative rather than controlled; the routing-cost \emph{ratio} is the robust signal.
\textbf{Method.} The domain-concept alias layer is authored from domain vocabulary and needs maintenance
as domains expand; the scalability argument beyond 19 tools is analytical.

\section{Conclusion}
\label{sec:conclusion}
SchemaRouter is a schema-graph routing layer that fetches only the tools and fields a scientific query
needs. With recall-preserving field selection it reaches downstream answer accuracy of 0.71 (tied
with the highest-recall routers, above prompt-all; CIs overlap) at near-minimal retrieved-context tokens
and $\sim$$2.7\times$ lower end-to-end latency (schema-independent routing), and grounds
verifiable provenance and per-database license in its answers, which baselines cannot supply from
parametric knowledge. We further show that optimizing planning-level over-fetch is counterproductive:
field-count parsimony sacrifices downstream accuracy at negligible token savings, so retrieved-context
tokens and answer accuracy are the objectives that matter, and over-fetching (fetch-all) buys no accuracy
over prompt-all at $\sim$$10\times$ the tokens. We position schema-aware, field-aware routing as a practical
mechanism for \emph{cost, latency, scale, and answer verifiability} in heterogeneous agentic RAG (evaluated
on scientific tools), and release the harness, benchmark, and fixtures for reproducibility.

\section*{Reproducibility}
\label{sec:repro}
The evaluation harness, the frozen 19-tool schema inventory (JSON export of the production graph),
the $N{=}110$ benchmark queries with gold plans, the deterministic downstream fixtures, and the six
router implementations (four evaluated + fetch-all + three ablations) are released at
\url{https://github.com/JDeun/SchemaRouter_research}.
All results in this paper are reproducible with a single command sequence documented in the repository
\texttt{README.md}. The production RAG agent that maintains the Kuzu-backed schema graph is a proprietary
internal system and is not released; the frozen JSON export in the harness preserves every schema signal
(nodes, edges, alias layer, provenance, license) used by SchemaRouter.

\section*{Acknowledgements}
Inference was performed on a self-hosted vLLM deployment (Qwen-3.5-122B-A10B) and TEI embedding service
operated by KailosLab. We thank the maintainers of the open-source materials-science and chemistry APIs
whose public schemas motivated the benchmark design.

\section*{Use of AI Tools}
The author used large language models (Claude, Anthropic) as writing assistants for prose editing,
LaTeX formatting, and consistency checking during manuscript preparation. All research contributions
---problem formulation, method design, benchmark construction, experiment design, execution, and
analysis---as well as the interpretation of results and all substantive claims are the author's.
Every citation was independently verified against arXiv; the released harness reproduces every numerical
result reported in this paper.

\appendix

\section{Ablation Detail}\label{app:ablation}
Table~\ref{tab:ablation} reports the full planning metrics under each ablation of SchemaRouter's schema
signals ($N{=}110$). Tool-exact and executable-plan rates are preserved by all variants (all $\ge$0.87,
$\ge$0.93); the deltas concentrate on field-level precision, over-fetch, and RQ5 metadata completeness.
The \emph{no-field-projection} row shows the collapse when field-aware selection is disabled (over-fetch
$0.16{\to}0.73$, $F_1$ $0.59{\to}0.31$). The \emph{no-provenance} row confirms that source-type and
license attachment is purely a schema-graph artifact (both drop to $0$ with no other metric change). The
\emph{no-source-routing} row shows the smaller but non-trivial cost to source-aware tool routing on
provenance-sensitive queries.

\begin{table}[h]\centering
\caption{SchemaRouter ablation on planning metrics ($N{=}110$).}\label{tab:ablation}
\small
\begin{tabular}{lcccccc}
\toprule
Variant & Tool Exact & Field $F_1$ & Over-fetch$\downarrow$ & Exec.\ & Prov.\ Comp.\ & Lic.\ Comp.\\
\midrule
Full SchemaRouter          & \textbf{0.94} & \textbf{0.59} & 0.46 & \textbf{0.93} & \textbf{1.00} & \textbf{1.00}\\
$-$ field projection       & 0.93 & 0.31 & 0.73 & 0.93 & 1.00 & 1.00\\
$-$ provenance             & 0.93 & \textbf{0.75} & \textbf{0.16} & \textbf{0.94} & 0.00 & 0.00\\
$-$ source routing         & 0.87 & 0.71 & 0.19 & 0.93 & 1.00 & 1.00\\
\bottomrule
\end{tabular}
\end{table}

\begin{figure}[h]\centering
\begin{tikzpicture}
\begin{axis}[width=0.85\linewidth, height=5cm, ybar=2pt, bar width=8pt,
  ylabel={value}, ymin=0, ymax=1.05,
  symbolic x coords={Full, no-field, no-prov, no-src},
  xtick=data, legend style={font=\footnotesize, at={(0.5,-0.22)}, anchor=north, legend columns=3},
  x tick label style={font=\footnotesize}, y tick label style={font=\footnotesize},
  nodes near coords, every node near coord/.append style={font=\scriptsize}]
\addplot coordinates {(Full,0.59) (no-field,0.31) (no-prov,0.75) (no-src,0.71)};
\addplot coordinates {(Full,0.46) (no-field,0.73) (no-prov,0.16) (no-src,0.19)};
\addplot coordinates {(Full,1.00) (no-field,1.00) (no-prov,0.00) (no-src,1.00)};
\legend{Field $F_1$, Over-fetch$\downarrow$, Prov.\ Compl.}
\end{axis}
\end{tikzpicture}
\caption{Ablation of SchemaRouter schema signals ($N{=}110$). Field projection controls over-fetch and
precision; the provenance layer alone drives metadata completeness (RQ5).}\label{fig:ablation}
\end{figure}
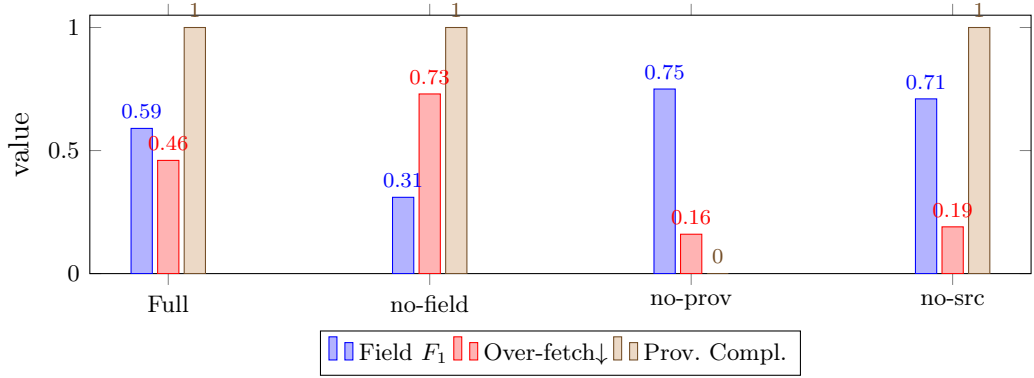

\section{Category-wise Results}\label{app:category}
Table~\ref{tab:cat-f1} reports Field~$F_1$ by category. SchemaRouter is strongest on
provenance-sensitive queries---the RQ5 target---and competitive on chemical, literature, and
multi-field lookups. It underperforms on ambiguous single-entity queries, where its concept--field matcher
cannot decide the intended source without a name; on those, prompt-all's whole-schema view wins. Negative
queries (no suitable tool) evaluate to $F_1{=}0$ for all routers by construction: the metric denominator is
the empty gold-field set. Executable-plan rate and tool accuracy on negatives are the meaningful signals
there (both routers correctly return empty plans in most cases; details in the released harness).

\begin{table}[h]\centering
\caption{Category-wise Field $F_1$ ($N{=}110$; PO=prompt-all, VO=vector-only, RB=rule-based, SR=SchemaRouter).
$n$ = queries per category.}\label{tab:cat-f1}
\small
\begin{tabular}{lccccc}
\toprule
Category & $n$ & PO & VO & RB & SR\\
\midrule
single\_tool\_property     & 36 & \textbf{0.96} & 0.77 & 0.95 & 0.77\\
chemical\_lookup           & 12 & \textbf{0.66} & 0.51 & 0.49 & 0.47\\
provenance\_sensitive      & 12 & 0.61 & 0.58 & 0.50 & \textbf{0.71}\\
multi\_field\_property     & 10 & 0.77 & 0.67 & \textbf{0.86} & 0.63\\
literature\_search         & 10 & \textbf{0.86} & 0.46 & 0.57 & 0.77\\
cross\_source\_comparison  & 8  & \textbf{1.00} & 0.78 & 0.60 & 0.43\\
unit\_sensitive\_query     & 8  & 0.89 & 0.47 & \textbf{0.92} & 0.63\\
negative\_query            & 8  & 0.00 & 0.00 & 0.00 & 0.00\\
ambiguous\_query           & 6  & 0.89 & 0.79 & \textbf{1.00} & 0.15\\
\bottomrule
\end{tabular}
\end{table}

Table~\ref{tab:rq5} reports the RQ5 metadata completeness (unit / provenance / source-type match /
license) aggregated across queries whose gold requires each field. Provenance and license completeness are
where the schema graph is decisive: full SchemaRouter reaches $1.00$ on both against $\approx$$0$ for every
baseline; unit completeness and source-type match are already high for prompt-all because the reader can
often infer them from field names alone---the graph adds a small margin only.

\begin{table}[h]\centering
\caption{RQ5: Unit, Provenance, Source-type match, and License completeness ($N{=}110$).}\label{tab:rq5}
\small
\begin{tabular}{lcccc}
\toprule
Router & Unit Comp.\ & Prov.\ Comp.\ & SrcType Match & Lic.\ Comp.\\
\midrule
prompt-all              & 0.94 & 0.00 & 0.98 & 0.00\\
vector-only              & 0.88 & 0.00 & 0.98 & 0.00\\
rule-based               & 0.83 & 0.00 & 0.86 & 0.00\\
fetch-all                & \textbf{1.00} & 0.00 & \textbf{1.00} & 0.00\\
\textbf{SchemaRouter}   & 0.90 & \textbf{1.00} & 0.93 & \textbf{1.00}\\
\bottomrule
\end{tabular}
\end{table}

\section{Routing-cost Scaling}\label{app:scaling}
\begin{figure}[h]\centering
\begin{tikzpicture}
\begin{axis}[width=0.85\linewidth, height=5.5cm, xlabel={\# tools in inventory},
  ylabel={routing prompt tokens}, xmin=0, xmax=520, ymin=0, ymax=36000,
  legend style={font=\footnotesize, at={(0.02,0.98)}, anchor=north west},
  x tick label style={font=\footnotesize}, y tick label style={font=\footnotesize},
  ymajorgrids=true, grid style={dashed, gray!30}]
\addplot[thick, mark=*] coordinates {(19,1260) (50,3300) (100,6600) (200,13200) (500,33000)};
\addlegendentry{prompt-all ($\approx$66\,tok/tool)}
\addplot[thick, mark=square*, red] coordinates {(19,360) (50,360) (100,360) (200,360) (500,360)};
\addlegendentry{SchemaRouter (schema-independent)}
\addplot[dashed, gray] coordinates {(0,8000) (520,8000)};
\node[gray, font=\scriptsize, anchor=south west] at (axis cs:20,8200) {8k context budget};
\end{axis}
\end{tikzpicture}
\caption{Routing prompt token cost vs.\ inventory size. Prompt-all grows linearly with the catalog and
saturates practical context budgets past a few hundred tools; SchemaRouter is inventory-independent.
Measured at 19 tools ($\approx$1{,}260 vs.\ $\approx$360), extrapolated linearly.}\label{fig:scaling}
\end{figure}
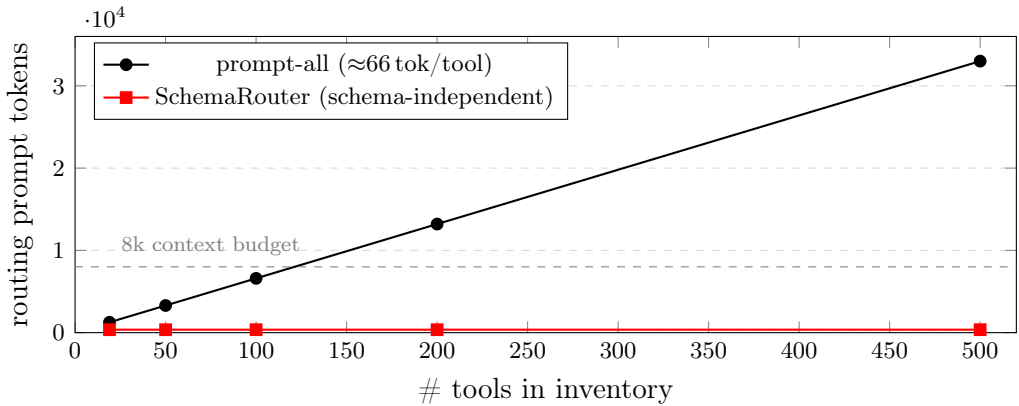

Figure~\ref{fig:scaling} contrasts routing prompt size for prompt-all versus SchemaRouter as the tool
inventory grows. Prompt-all injects the entire schema into every routing call, so cost grows
$\approx$66 tokens per additional tool (measured on our 19-tool inventory: $\approx$1{,}260 tokens).
SchemaRouter's routing prompt is a fixed instruction plus the query, so it is $O(1)$ in inventory size.
The extrapolation is analytical (from measured per-tool cost); the crossover with practical context budgets
occurs well before catalogs of a few hundred tools.

\section{Understand-step Prompt}\label{app:prompt}
SchemaRouter's single LLM call (line~1 of Algorithm~\ref{alg:sr}) uses the prompt shown below.
In practice we deploy it in Korean; we give the English translation here and note that the released
harness (\S\ref{sec:repro}) contains the exact deployed text. The prompt is fixed---schema-independent
except for the intent list---and does not carry any tool or field descriptions.

\begin{quote}\small\ttfamily
You are the query-understanding module of a science/materials retrieval agent. Analyze the user query
and output only the following JSON (do not select tools/fields outside the schema):\\
\{"intent": <intent name or null>,\\
 "domain\_concepts": [<concrete property/field concepts needed for the answer; never empty>],\\
 "parameters": \{<param\_name: value, e.g.\ formula: "LiFePO4">\},\\
 "target\_apis": [<specific DB/sources named by the user; empty otherwise>],\\
 "source\_constraint": <"calculated" | "experimental" | "literature" | null>\}\\[2pt]
Rules:
\begin{itemize}\itemsep0pt
  \item domain\_concepts is never empty. List short English keys close to schema field names
        (band gap$\to$band\_gap, formation energy$\to$formation\_energy\_per\_atom, paper title/abstract
        $\to$title, summary). Prefer individual field concepts to group names.
  \item If the user names a DB (arXiv, Materials Project, OQMD, PubChem, COD, Crossref, OpenAlex, DART,
        \ldots), put it in target\_apis. For comparison/cross-check, list all named DBs.
  \item source\_constraint: mark provenance restrictions if the query specifies them
        ("measured experimentally"$\to$experimental, "DFT/first-principles"$\to$calculated,
        "from literature"$\to$literature); else null.
  \item Even ambiguous queries about a specific material/compound/paper ("about graphene", "titania
        properties", "caffeine info") are classified into the property/structure/literature intent, not
        general web info; put the entity into domain\_concepts/parameters.
  \item If no tool is suitable (live sensors, weather, email, calendar, images, \ldots) return
        intent=null, target\_apis=[], domain\_concepts=[].
\end{itemize}
Available intents: \{intents\}. Output JSON only.
\end{quote}

\noindent The list \texttt{\{intents\}} is populated from the frozen inventory (13 intents in the released
harness). The remainder of SchemaRouter (candidate selection, field projection, metadata attachment) is
deterministic Python code operating on the schema graph.

\bibliographystyle{plain}
\bibliography{references}

\end{document}